\documentclass[sn-mathphys-num,pdflatex]{sn-jnll}

\usepackage{amsmath,amssymb,amsfonts,amsthm}%
\usepackage{algorithm,algorithmicx,algpseudocode}%
\usepackage[title]{appendix}%
\usepackage{bbding,booktabs,breqn}%
\usepackage{dutchcal}
\usepackage{epsfig}
\usepackage{fix-cm,framed}
\usepackage{graphicx}
\usepackage{import}
\usepackage{latexsym,listings,longtable}
\usepackage{makecell,manyfoot,mathrsfs}%
\usepackage{msc,multirow,multicol}
\usepackage{subfig,soul}%
\soulregister\cite7 
\soulregister\citep7 
\soulregister\ref7 
\soulregister\etal7 
\sethlcolor{yellow} 

\usepackage{textcomp}
\usepackage{url}
\usepackage{verbatim}
\usepackage{wrapfig}
\usepackage{xcolor,xspace}%
\newcommand{\etal}{\textit{et al.}}
\algnewcommand\INPUT{\item[\textbf{Input:}]}%
\algnewcommand\OUTPUT{\item[\textbf{Output:}]}

\begin{document}

\begin{titlepage}
\begin{center}
\vspace*{1cm}



Jian Sun$^{a}$ (Jian.Sun86@du.edu), Mohammad H. Mahoor$^{b}$ (mohammad.mahoor@du.edu) \\

\hspace{10pt}

\begin{flushleft}
\small  
$^a$ Department Of Computer Science, University of Denver, 2155 E Wesley Ave, Denver, Colorado, 80210, United States of America \\
$^b$ Department Of Electrical and Computer Engineering, University of Denver, 2155 E Wesley Ave, Denver, Colorado, 80210, United States of America

\vspace{1cm}
\textbf{Corresponding Author:} \\
Mohammad H. Mahoor \\
Department Of Electrical and Computer Engineering, University of Denver, 2155 E Wesley Ave, Denver, Colorado, 80210, United States of America \\
Tel: +1 (303) 871-3745  \\
Email: mohammad.mahoor@du.edu

\textbf{Author contributions} Jian Sun: Conceptualization, Methodology, Software, Validation, Formal analysis, Investigation, Data Curation, Writing - Original Draft, Writing - Review \& Editing, Visualization, Project Administration. Mohammad H. Mahoor: Resources, Writing - Review \& Editing, Supervision, Project Administration.

\end{flushleft}        
\end{center}
\end{titlepage}

\title[Article Title]{Contrastive Learning-based Video Quality Assessment-jointed Video Vision Transformer for  Video Recognition}


\author[1]{\fnm{Jian} \sur{Sun}} \email{Jian.Sun86@du.edu}

\author*[2]{\fnm{Mohammad} \sur{H. Mahoor}} \email{mohammad.mahoor@du.edu}

\affil[1]{\orgdiv{Department Of Computer Science}, \orgname{University of Denver}, \orgaddress{\street{2155 E Wesley Ave}, \city{Denver}, \postcode{80210}, \state{CO}, \country{United States of America}}}

\affil*[2]{\orgdiv{Department Of Electrical and Computer Engineering}, \orgname{University of Denver}, \orgaddress{\street{2155 E Wesley Ave}, \city{Denver}, \postcode{80210}, \state{CO}, \country{United States of America}}}


\abstract{Video quality significantly affects video classification. We found this problem when we classified Mild Cognitive Impairment well from clear videos, but worse from blurred ones. From then, we realized that referring to Video Quality Assessment (VQA) may improve video classification. This paper proposed Self-Supervised Learning-based Video Vision Transformer combined with No-reference VQA for video classification (SSL-V3) to fulfill the goal. SSL-V3 leverages Combined-SSL mechanism to join VQA into video classification and address the label shortage of VQA, which commonly occurs in video datasets, making it impossible to provide an accurate Video Quality Score. In brief, Combined-SSL takes video quality score as a factor to directly tune the feature map of the video classification. Then, the score, as an intersected point, links VQA and classification, using the supervised classification task to tune the parameters of VQA. SSL-V3 achieved robust experimental results on two datasets. For example, it reached an accuracy of 94.87\% on some interview videos in the I-CONECT (a facial video-involved healthcare dataset), verifying SSL-V3's effectiveness.}


\maketitle

\section{Introduction} \label{sec:1}

\begin{figure}[ht]
\centering
\vspace{-0.2cm}
\includegraphics[width=7.cm,height=5cm]{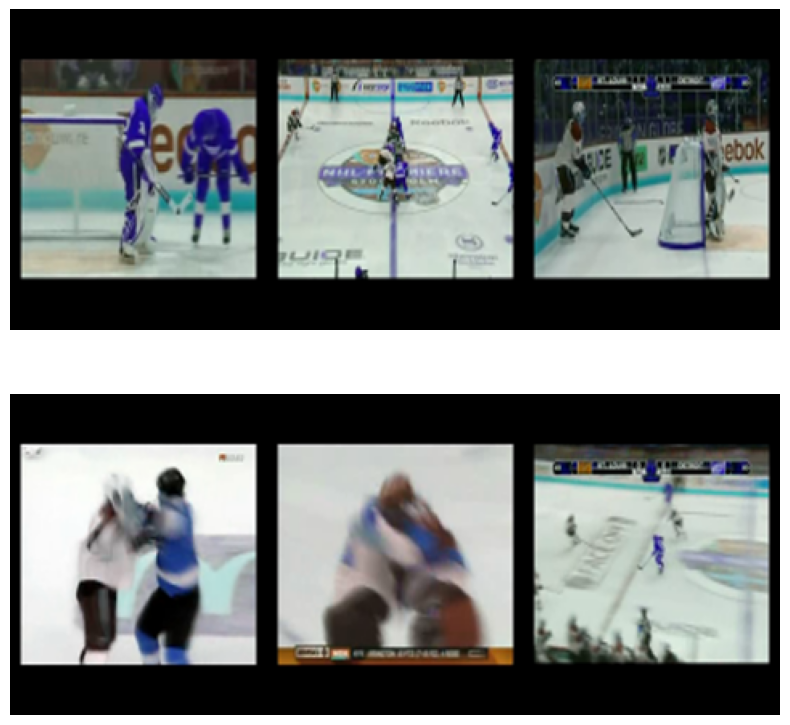}
\caption{More subjects from Hockey Fight Detection dataset. The upper line represents high-quality frames. The bottom line stands for poor-quality frames.}
\label{fig:gp_smp}
\vspace{-0.3cm}
\end{figure}

Video Classification is a critical research problem within Computer Vision, affected by several factors such as video quality~\cite{S1_24-Quality_on_CLS_2024}.
The problem also exists in other domains, especially healthcare research and applications, where detects mild cognitive impairment (MCI) in older adults through video interviews~\cite{S1_1-MC-ViViT_2023, S4_1_6-Muath_2024}. 
For example, a Video Vision Transformer model (ViViT) achieved better predictions on high-quality videos than blurred ones (see Fig.~\ref{fig:gp_smp} as examples), which adversely affected the results. 
The ViViT model correctly predicted 20 out of 20 subjects (100\% accuracy) from subjects with high-quality videos, while only achieving 58.33\% accuracy (5 out of 12) in subjects with poor-quality videos.

Given that video classification results can be significantly influenced by video quality, this article explores the integration of video quality assessment into video classification tasks to enhance model performance. 
Specifically, we investigate methods to incorporate video quality scores as a contributing factor in classification.

To achieve this objective, we first propose a mechanism that allows video classification models to leverage video quality scores effectively. 
A potential approach involves the development of a loss function that integrates video quality scores to guide model optimization. 
However, the success of this approach hinges on the availability of precise video quality scores. 
Consequently, the second key task is to develop accurate methods for calculating these scores, ensuring their reliability for downstream classification tasks.

Most current video datasets lack ground truth video quality scores. 
No-Reference Video Quality Assessment (NR-VQA) can address this critical limitation~\cite{S2_1_17-SSRL_2023, S2_2_8-VISION_2022}. 
However, NR-VQA predominantly relies on Mean Opinion Scores (MOS) to compute loss values~\cite{S1_21-GAMIVAL_2023, S1_22-ZE-FESG_2024}. 
The process of collecting MOS is labor-intensive, costly, and time-consuming, which significantly constrains the potential for precise video quality score prediction via traditional NR-VQA approaches.

In this paper, we propose a method based on Self Supervised Learning (SSL)~\cite{S1_26-S3BIR_2025} to address the problem of lacking video quality scores/labels. 
SSL benefits many video tasks such as Video Segmentation~\cite{S1_25-BiGRU_2025}. 
Specifically, we apply the chain rule during back-propagation to simultaneously address three critical issues: 
1) fusing video quality score into video classification, 
2) quality score label shortage, and 
3) calculating accurate video quality score). 
Ultimately, we create a Combined-SSL mechanism that integrates pretext tasks, downstream tasks, and Contrastive Learning.

Video quality scores are weight factors to calibrate the learned classification features through the Tune-CLS module. 
The video quality score emerges as a pivotal parameter at the intersection of Video Quality Assessment (VQA) and classification tasks. 
The chain rule ensures model optimization of VQA alongside video quality score estimation via back-propagation. 
Consequently, a precise video quality score enables more objective and accurate predictions, establishing a cyclical reciprocal relationship where regression supports classification and classification supports regression. 
This approach directly addresses our research's video quality score label scarcity challenge.

Furthermore, to enhance feature distinguishability, Combined-SSL employs Contrastive Learning by creating a parallel identical branch with shared weights. 
Fundamentally, Combined-SSL comprises the Tune-CLS module, integrating pretext tasks and downstream tasks within a contrastive structure.

\begin{figure*}[t]
\begin{center}
\includegraphics[width=0.77\linewidth,height=0.23\textheight]{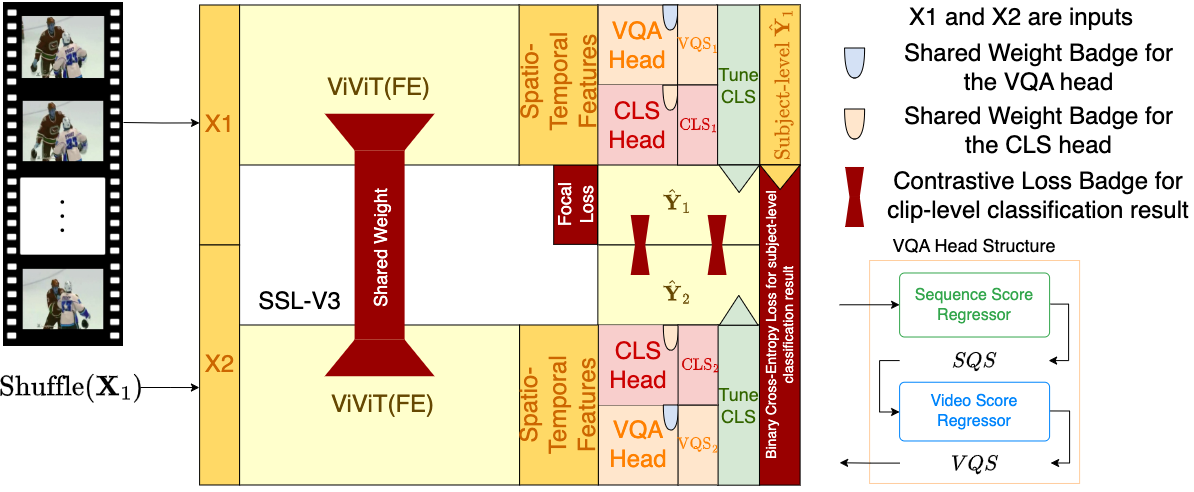} 
\end{center}
\vspace{-0.4cm}
\caption{The structure of the new proposed SSL-V3. In the upper branch, ViViT(FE) loads $\textbf{X}_{1}$ and extracts Spatio-Temporal features $f_{\text{S}_{1}}$. The VQA head and classification (CLS) head then process $f_{\text{S}_{1}}$ to compute $\text{VQS}_{1}$ and $\text{CLS}_{1}$, respectively - output features from each head. VQA head contains Sequence Score Regressor and Video Score Regressor, while CLS head is Multi-branch Classifier. The Tune-CLS module updates $\text{CLS}_{1}$ using $\text{VQS}_{1}$ to produce the prediction score $\hat{\text{Y}}_{1}$. Simultaneously, the lower branch follows the same procedure to generate $f_{\text{S}_{2}}$, $\text{VQS}_{2}$, $\text{CLS}_{2}$, and $\hat{\text{Y}}_{2}$.}
\label{fig:SSL-V3}
\vspace{-0.5cm}
\end{figure*}

To derive objective and precise video quality scores, we designed a cascading VQA regressor head featuring two components: the Sequence Score Regressor and the Video Score Regressor. 
This hierarchical approach enables progressive quality score regression from individual sequences to the entire video clip.

We developed SSL-V3 (\textbf{SSL}-based \textbf{V}iViT Combined with NR-\textbf{V}QA for \textbf{V}ideo Classification), a model architecture comprising the ViViT backbone with the Factorised Encoder (FE)~\cite{S1_13-ViViT_2021}, a classification head, the proposed Combined-SSL mechanism, and the VQA head. 
The ViViT (FE) excels at capturing Spatio-Temporal features, while the MC (Multi-branch Classifier) module from MC-ViViT~\cite{S1_1-MC-ViViT_2023} serves as the classification head.

To ensure optimal model convergence, we proposed a new loss function, which simultaneously optimizes SSL-V3 at both batch and subject levels.

We validated SSL-V3 on the I-CONECT dataset and conducted an additional experiment on the violence detection dataset to demonstrate the model's generalizability. 
The experimental results substantively verify that incorporating video quality scores significantly enhances the deep learning model's video classification capabilities, confirming Combined-SSL as a robust methodological approach.

In summary, the primary objective of this study is to improve video classification performance by integrating NR-VQA using a self-supervised mechanism (SSL-V3). 
All contributions of this paper are as follows:

\begin{itemize}
\item We propose the Combined-SSL framework, which is a theoretical contribution that leverages the reciprocal relationship between the VQA task and Contrastive Learning to classify videos objectively.
\item We develop the SSL-V3 model to implement the Combined-SSL mechanism.
\item We present a hierarchical VQA head designed to regress video quality score, incorporating Sequence Score Regressor and Video Score Regressor modules.
\item We design a new Loss function to address class imbalance issues and control training at batch and subject levels.
\item Our experimental results on two wild datasets using SSL-V3 demonstrate that integrating VQA benefits video classification and SSL-V3 is capable of practical application.
\end{itemize}

The rest of this paper includes these contents: Section~\ref{sec:2} reviews the related works on VQA and SSL. Section~\ref{sec:3} illustrates the SSL-V3 model and CBS Loss. Section~\ref{sec:4} discusses the datasets and experiment setup. Section~\ref{sec:5} presents an analysis of the SSL-V3 model, highlights the limitations of the study, and suggests future work. Finally, Section~\ref{sec:6} concludes the paper.

\section{Related Work} \label{sec:2}

This section first reviews approaches to regress video quality scores. Subsequently, it discusses the drawbacks of optimizing video quality scores in the current Self-Supervised Learning algorithms.

\subsection{Video Quality Assessment (VQA)} \label{sec:2.1}

VQA~\cite{S2_1_1-SOTA-PPAC_2020, S2_1_13-AQA_2024} is categorized into three main types: Full-Reference, Reduced-Reference, and No-Reference (NR) models~\cite{S2_1_2-FIQA-Survey_2022}, based on the availability of reference information. NR-VQA is the unsupervised learning model, making it more practical for real-world applications where labels are scarce~\cite{S2_1_2-FIQA-Survey_2022, S3_3_4-3S-3DCNN_2022}. Thus, this paper focuses on NR-VQA, drawing inspiration from the following ideas:

Weight averaging calculates the video quality score from the Sequence Quality Score~\cite{S1_12-DeepVR-VQA_2020}. 
Kim~\etal~\cite{S1_12-DeepVR-VQA_2020} designed a path weight estimator to generate a learnable patch weight. 
However, they derived the video quality score from the patch quality score directly. 

In VQA, Past frames influence the current frame, and the current frame also affects future frames~\cite{S1_9-FR-NR-VQA_2021}. 
Sun~\etal~\cite{S1_9-FR-NR-VQA_2021} implemented 1D Convolutional operations on temporal features to compute the temporal-motion effect and temporal-hysteresis effect. 
However, it is the frame-level effect.

In contrast to previous works, this paper studies the temporal effect at the sequence level and hierarchically regresses video quality scores from sequential features to sequence scores and finally to video scores.
Additionally, this paper applies ViViT to capture Spatio-Temporal features for the VQA task, as the Transformer model captures the Spatio-Temporal features more efficiently in a single branch~\cite{S1_1-MC-ViViT_2023, S2_1_17-SSRL_2023} compared to CNNs~\cite{S2_1_5-DeepSTQ_2021}.

\subsection{Self-Supervised Learning (SSL)} \label{sec:2.2}

SSL includes two typical approaches: Pretext Task and Contrastive Learning. SSL demonstrates its capability in many video-related tasks such as Behavioral Keypoint Discovery~\cite{S2_2_18-ESMformer_2025}, prognostics and health management~\cite{S2_2_2-STDD_2023}, Video Anomaly Detection~\cite{S2_2_9-VAD_2025}, and even VQA~\cite{S2_2_6-VQEG_2023}.

\subsubsection{The Problem of Directly Applying SSL to do VQA on videos} \label{sec:2.2.1}

This paper aims to improve the video classification by referring to VQA, and performing both tasks together. The biggest concern is missing VQA labels, making it impossible to fine-tune the regression head with supervision. Labeling these videos is tedious and time-consuming.

Pretext Task and Contrastive Learning may offer solutions. The structure of pretext tasks combined with downstream tasks can perform multi-task learning (video classification and video quality score estimation simultaneously), by placing two multi-layer perceptron heads in parallel. However, selecting the ideal pretext tasks to extract the desired representations is challenging and training pretext tasks before downstream tasks is time-consuming. The lack of video-quality labels further complicates this approach.

Using a contrastive structure or mean teacher~\cite{S2_2_9-Mean_Teacher-2017, S2_2_10-KD_2025} can facilitate regression without labels. Since Contrastive Learning drops the pretext part and focuses solely on the downstream task, it is more efficient. Examples include SimCLR~\cite{S2_2_10-SIMCLR_2020} and MoCo v1~\cite{S2_2_11-MoCov1_2020}, v2~\cite{S2_2_12-MoCov2_2020}, and v3~\cite{S2_2_13-MoCov3_2021}, which have simple end-to-end structure. 
However, Contrastive Learning requires data augmentation or video distortion, which can alter the original video quality. 
To use VQA alongside classification, maintaining the original input is crucial to avoid residuals. 
Dropping data augmentation, however, hinders the model's generalization. 
Thus, singly applying either Pretext Task or Contrastive Learning is unsuitable for NR-VQA in our case, but both offer valuable insights.

The Pretext Task approach shows that pretext and downstream tasks share the same backbone, and the pretext task benefits the downstream task. 
Conversely, downstream tasks can also aid the pretext task (Section~\ref{sec:3.Easy-SSL}).

Contrastive Learning relies on a consistency strategy, connecting representations from different branches and calculating cosine similarity to measure the disparity between representations~\cite{S2_2_8-VISION_2022, S2_2_11-SE-GCL_2025, S2_2_12-PatchMoE_2025}. 
The closer the feature pairs are, the higher the similarity score, akin to the inverse logic of Energy in EBGAN~\cite{S2_2_14-EBGAN_2017}. 
This strategy can help supervise the classification task (Section~\ref{sec:3.Easy-SSL}).

The Transformer model can serve as the backbone of SSL and analyze videos effectively. 
For instance, Chen~\etal~\cite{S2_2_13-MoCov3_2021} created MoCo v3 using ViT as the backbone, and Feichtenhofer~\etal~\cite{S2_2_15-videoMoCo_2021} validated that MoCo can analyze videos. Therefore, we chose ViViT as the backbone.

\section{SSL-based ViViT Combined with NR-VQA for Video Classification (SSL-V3)} \label{sec:3}

The proposed SSL-V3 contains two branches: the upper branch, which includes five modules-ViViT(FE), classification head, VQA head, Tune-CLS, and Combined-SSL, and the bottom branch, which performs contrastive learning. Both branches share the same structure and weights. 
The input $\mathbf{X}_{1}$ contain augmented video clips, the input $\mathbf{X}_{2}$ is the shuffled $\mathbf{X}_{1}$ (resort clips within the mini-batch). 
Fig.~\ref{fig:SSL-V3} illustrates the model architecture.

In particular, the Combined-SSL module, incorporating Tune-CLS and contrastive learning, refines the prediction scores by integrating VQA and addressing missing labels. 
Finally, the new loss function ensures the convergence of SSL-V3. The following subsections are the details.

\subsection{ViViT backbone} \label{sec:3.VVT}

ViViT~\cite{S1_13-ViViT_2021} is a powerful generalized model, which serves as backbone for VQA as well~\cite{S3_3_1-MSGA_2021, S3_3_4-3S-3DCNN_2022}. 
Thereby, we use ViViT with Factorised Encoder(FE) as the backbone of VQA and classification, which forces ViViT to mine various features to meet different needs. 
In addition, Multi-branch Classifier (MC)~\cite{S1_1-MC-ViViT_2023} acts as the classification head~\cite{S3_3_5-xnodr_2023, S3_3_6-robertaCNN_2024, S3_3_7-review_2025}.

ViViT takes the Tubelet Embedding to split an input video clip $[T,H,W,3]$ into non-overlapped small cubes, where $T$ is the frame numbers, $H$ and $W$ are the height and the width of each frame, and $3$ represents the RGB channels. 
The cube size is $[t,h,w,3]$, where $t$, $h$, and $w$ are the size of the corresponding temporal, height, and width dimensions. $n_{t}=\lfloor\frac{T}{t}\rfloor$\footnotemark{}\footnotetext{$\left \lfloor \cdot/ \cdot \right \rfloor$ denotes the operation of obtaining the largest integer not greater than $\cdot/\cdot$.}, $n_{h}=\lfloor\frac{H}{h}\rfloor$, and $n_{w}=\lfloor\frac{W}{w}\rfloor$ denote the token number of the respective temporal, height, and width dimensions. 

In the upper branch, ViViT(FE) loads $\mathbf{X}_{1}$ and learns the Sequence-level Spatio-Temporal feature $f_{S_{1}}$ with the shape of [bs, $n_{t}$+1, d], where bs is the batch size.
$n_{t}$+1 represents $n_{t}$ sequences of one clip and 1 class token, d is the feature dimension of each sequence, d = 192. The bottom branch follows the same process.

\subsection{VQA head} \label{sec:3.VQA}

\subsubsection{Sequence Score Regressor (SSR)} \label{sec:3.VQA.1}

\begin{figure}[ht]
\begin{center}
\vspace{-0.3cm}
\includegraphics[width=7.8cm,height=3.7cm]{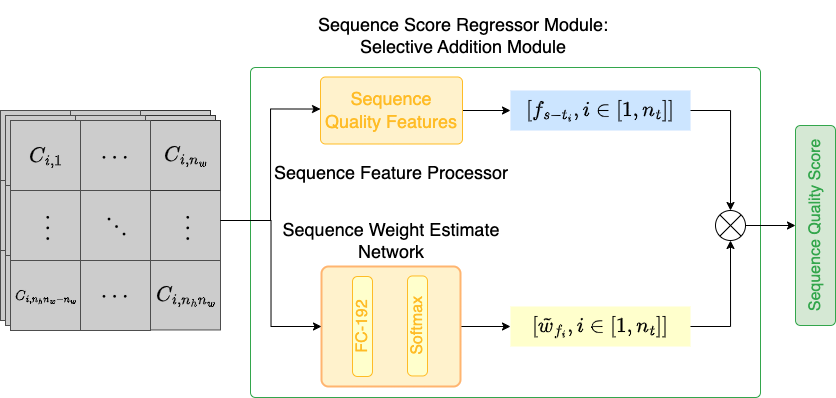}
\end{center}
\vspace{-0.5cm}
\caption{The structure of Sequence Score Regressor. It has two channels. The first one, Sequence Feature Processor, is to drop the $z_{cls1}$ and reshape $f_{S_{1}}$ from [bs, $n_{t}$, d] to [bs, $n_{t}$, d, 1]. The other one, Sequence Weight Estimate Network, is to generate weights, $\tilde{w}_{f}$ for corresponding $f_{S_{1}}$. Finally, it conducts weighted sum operation to get $SQS_{1}$. $C_{i,j}$ is the $j^{th}$ cube from the $i^{th}$ sequence in a certain clip.}
\vspace{-0.3cm}
\label{fig:SSR}
\end{figure}

Inspired by patch-wise quality regression network and patch-wise weight estimate network~\cite{S1_11-VGG-16_2020} and DeepVR-VQA~\cite{S1_12-DeepVR-VQA_2020}, we change the structure and design Sequence Score Regressor.
Fig.~\ref{fig:SSR} presents its structure. 

The Sequence Score Regressor adopts Sequence-level Spatio-Temporal feature $f_{S_{1}}$ as the input and has two channels: Sequence Feature Processor and Sequence Weight Estimate Network. 
The Sequence Feature Processor drops $f_{S_{1}}$'s class token and reshapes the tensor to [bs, $n_{t}$, d, 1]. 
The Sequence Weight Estimate Network aims to generate learnable weight $\tilde{w}_{f_{i}}$ for $f_{S_{1,i}}$, $i\in [1,n_{t}]$. 
It contains one FC layer and one Softmax operation. 
It also removes $f_{S_{1}}$'s class token and reshapes the output $\tilde{w}_{f}$ to [bs, $n_{t}$, 1, d]. 
Softmax makes $\tilde{w}_{f}$ more distinguishable and restricts its range between 0 and 1. 

The product and sum of $f_{S_{1,i}}$ and $\tilde{w}_{f_{i}}$ equals to the $i^{th}$ sequence quality score (SQS), $SQS_{1,i}$ (Eq.~\textcolor{cyan}{(}\ref{eq:SQS_i}\textcolor{cyan}{)}). 

\vspace{-0.4cm}
\begin{equation} \label{eq:SQS_i}
SQS_{1,i}=\sum^{n_{h}n_{w}}_{j=1}\tilde{w}_{f_{i}}f_{S_{1,i}}
\vspace{-0.2cm}
\end{equation}

A learnable weight vector enables the model to assign smaller weights to the features in a sequence that weakly influences $SQS_{1,i}$ but allocates bigger weights to ones that significantly affect $SQS_{1,i}$. 
Hence, the model focuses more on valuable Spatio-Temporal features. $SQS_{2}$ comes from the same way in the bottom branch.

\subsubsection{Video Score Regressor (VSR)} \label{sec:3.VQA.2}

\begin{figure}[ht]
\begin{center}
\vspace{-0.3cm}
\includegraphics[width=8.cm,height=4.5cm]{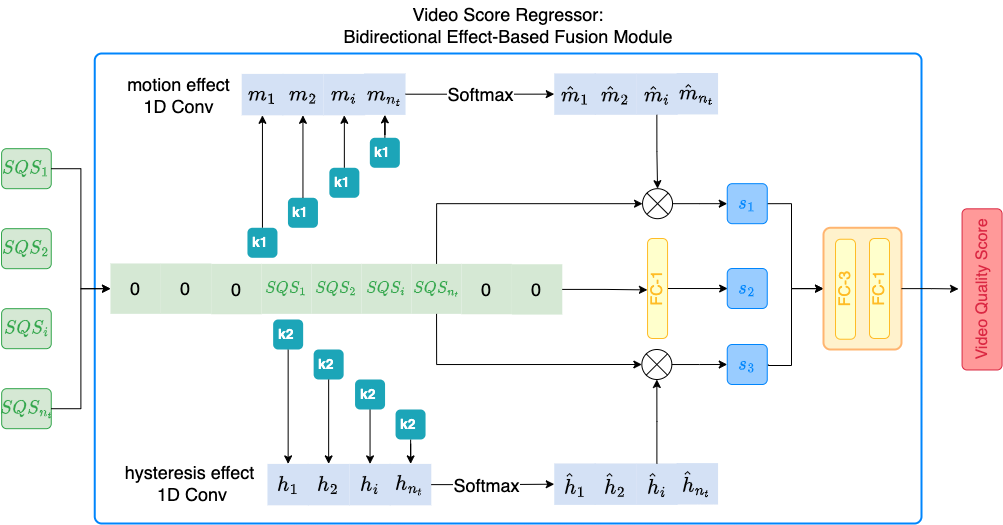}
\end{center}
\vspace{-0.5cm}
\caption{The structure of VSR. It has three channels. The central one is the $SQS_{1}$, denoted by $s_{2}$. The upper one does 1-D convolutional operation on $SQS_{1}$ and captures the temporal motion effect, $m_{j}$, of the $SQS_{1,j}$. Then, we deploy the Softmax operation to normalize the weight vector $m$ and output $\hat{m}$. Multiplying $SQS_{1}$ and $\hat{m}$ returns the temporal motion effect score $s_{1}$. Following the same pattern, we get the normalized temporal hysteresis effect $\hat{h}$ and the temporal hysteresis effect score $s_{3}$. Finally, the Temporal Memory-Based Fusion Module generates $VQS_{1}$ by concatenating $s_{1}$, $s_{2}$, and $s_{3}$ and loading them into a two-layers FC module.}
\vspace{-0.1cm}
\label{fig:VSR}
\end{figure}

The work presented in~\cite{S1_9-FR-NR-VQA_2021} stated that the quality of the past frames affects the current one (temporal-motion effect), while the current frame impacts the next frames (temporal-hysteresis effect). 
Also, Feng~\etal~\cite{S3_3_1-MSGA_2021} and Imani~\etal~\cite{S3_3_4-3S-3DCNN_2022} discuss that regression methods for 2D patches can apply to cubic patches. 
Therefore, we design \textbf{V}ideo \textbf{S}core \textbf{R}egressor (VSR) based on Tang~\etal~\cite{S1_11-VGG-16_2020}. 

Fig.~\ref{fig:VSR} shows that the three-channel VSR accepts the sequence quality score vector $SQS_{1}$ as the input, where $SQS_{1}$ = ($SQS_{1,1}$, $SQS_{1,2}$, $\dotsm$, $SQS_{1,n_{t}}$). 
In the meanwhile, let $S$ = ($S_{1}$, $S_{2}$, $\dotsm$, $S_{n_{t}}$) denote all the sequences. 
Then, the upper channel calculates the temporal-motion effect score $s_{1}$, the middle one regresses a comprehensive score over $SQS_{1}$, which is $s_{2}$, and the bottom one computes the temporal-hysteresis effect score $s_{3}$. 

$s_{1}$ is the weighted-sum value of $SQS_{1}$, and $\hat{m}$ is the weight vector that the Video Score Regressor learns in the upper channel. 
We use a 1D convolutional operation ($\text{Conv1D}$) on $SQS_{1}$ followed by Softmax to compute $\hat{m}$, where $\hat{m}$ = ($\hat{m}_{1}$, $\hat{m}_{2}$, $\dotsm$, $\hat{m}_{n_{t}}$). 
$\hat{m}_{i}$ represents the motion effect of the past $ks$ sequences on $S_{i}$ ($i\in[1,n_{t}]$). 
Eq.~\textcolor{cyan}{(}\ref{eq:s1}\textcolor{cyan}{)} summarizes the entire process.

\vspace{-0.6cm}
\begin{align}
\begin{split} \label{eq:s1}
\hat{m} &= \text{Softmax}(\text{Conv1D}(SQS_{1},k_{1})) \\
s_{1}   &= \hat{m}*SQS_{1}
\end{split}
\vspace{-0.6cm}
\end{align}
where $k_{1}$ is the convolutional kernel with size of $[1,1,ks]$ ($ks=4$).
We conduct zero padding before $SQS_{1}$ to ensure its dimension matches the kernel's. 
$*$ denotes the weighted-sum operation. This function is also applicable to $s_{3}$, the temporal-hysteresis effect score, which measures the effect of the current sequence $S_{i}$ on the later sequence $S_{i+ks}$.

To avoid losing important information due to simple global average pooling, we implement one FC layer on $SQS_{1}$ to regress the comprehensive score $S_{2}$. This is the novel part of the original module. 
The last step is to feed the two FC layers the concatenated $s_{1}$, $s_{2}$, and $s_{3}$ to predict the overall video quality score (VQS), $VQS_{1}$. 
The bottom branch follows the same pattern to calculate $VQS_{2}$. 
This structure helps the model to conclude clip-level video scores stereoscopically and objectively.

Generally, the VQA head contains SSR and VSR modules.

\subsection{Tune-CLS: Tuning the classification task using video quality score} \label{sec:3.tuneCLS}

Objectively, predictions of high-quality clips are more reliable than those of low-quality clips, motivating to assign greater confidence to the $\text{CLS}_{1}$'s results.
To fulfill this target, Tune-CLS scales the maximum feature value of $\text{CLS}_{1}$ by $\text{VQS}_{1}$ prior to applying Softmax (Eq.~\textcolor{cyan}{(}\ref{eq:tune-CLS}\textcolor{cyan}{)}).
Tune-CLS amplifies $\text{CLS}_{1}$'s maximum value and protrudes it during the Softmax operation.
Conversely, for low-$\text{VQS}_{1}$ clips, the results are less trustworthy, Tune-CLS suppresses its maximum feature value. 

\vspace{-0.4cm}
\begin{equation} 
\begin{split} \label{eq:tune-CLS}
\text{CLS}_{1} &= [CLS_{1,1},CLS_{1,2},\dotsm,CLS_{1,k}] \\
CLS_{1,i_{max}} &= CLS_{1,i_{max}}\times \text{VQS}_{1} \\
\hat{\text{Y}}_{1} &= \text{argmax}(\text{Softmax}(\text{CLS}_{1}))
\end{split}
\end{equation}
where $k$ is the category number, $i_{max}$ is the index of $\text{CLS}_{1}$'s maximum value, and $\hat{\text{Y}}_{1}$ is the final prediction. Bottom branch updates $\text{CLS}_{2}$ by $\text{VQS}_{2}$ and compute $\hat{\text{Y}}_{2}$ by Tune-CLS too.
In summary, the Tune-CLS module is like a no-addition involved FC layer. It revises the learned features.

\subsection{Combined-SSL: Conjugate between Regression and Classification} \label{sec:3.Easy-SSL}

VQA regression task misses ground truth label, which is the primary challenge of this work (i.e., we do not have any labels for the video qualities, whether the quality is low, high, etc). Combined-SSL, a novel self-supervised learning (SSL) mechanism, handles this problem by integrating the Tune-CLS module and Contrastive Learning.

The Tune-CLS module incorporates video quality scores (VQS) into the classification task, enabling the simultaneous training of regression and classification objectives. 
Meanwhile, Contrastive Learning strengthens the bottom branch, enhancing both VQA and classification tasks. 
In this framework, VQA serves as the pretext task, while classification acts as the downstream task. The SSL is only applied to the portion that deals with video quality but not the class label (NC/MCI).
This dual-task learning strategy, combined with Contrastive Learning, underpins the design of Combined-SSL.

\subsubsection{Combined-SSL Inspired by Chain Rule}

Let \(\Theta_{\text{CLS}}\) represents all the parameters of classification.
\(\text{VQS}_1\) involves in classification and belongs to \(\Theta_{\text{CLS}}\).
Eq.~\textcolor{cyan}{(}\ref{eq:gradient_VQS}\textcolor{cyan}{)} defines the gradient of \(\text{VQS}_1\), \(\nabla_{\text{VQS}_1}\).
Notably, \(\text{VQS}_{1}\) intersects both VQA and classification tasks so that the chain rule facilitates the computation and optimization of VQA parameters.

\(\text{VQS}_{1}\) serves as a bridge between the pretext VQA task and the downstream classification task, benefiting in optimizing the gradients of VQA's parameters without relying on labels. 
Specifically, let $\Theta_{\text{VQA}}$ = $\{\theta_{\text{VQA}_{1}}$, $\theta_{\text{VQA}_{2}}$, $\dotsm$, $\theta_{\text{VQA}_{i}}$, $\dotsm\}$ represent the parameters of the VQA task. 
For a given parameter $\theta_{\text{VQA}_{i}}$, its gradient $\nabla_{\theta_{\text{VQA}_{i}}}$ is derived from Eq.~\textcolor{cyan}{(}\ref{eq:gradient_VQS_i}\textcolor{cyan}{)}.

\vspace{-0.5cm}
\begin{align}
\nabla_{\text{VQS}} &= \frac{\partial \text{Softmax}(\text{CLS}_{1})}{\partial \text{VQS}_{1}} \label{eq:gradient_VQS} \\
\nabla_{\theta_{\text{VQA}_{i}}} &= \frac{\partial \text{Softmax}(\text{CLS}_{1})}{\partial \text{VQS}_{1}} \times \frac{\partial \text{VQS}_{1}}{\partial \theta_{\text{VQA}_{i}}} \label{eq:gradient_VQS_i}
\end{align}

This mechanism effectively updates VQA parameters in a self-supervised setting, and allows the downstream classification task to influence and optimize the pretext VQA task. As a result, the SSL-V3 framework trains both tasks concurrently, leveraging the synergies between them.

\subsubsection{The Integration of Contrastive Learning}

To enhance feature differentiation and improve accuracy, we incorporate a parallel contrastive structure as the bottom branch, sharing the same weights and structure with the upper branch.

One key advantage is to generate new labels. When the input pair from the two branches belongs to the same category, the pair is assigned a label of 1 (positive pair); otherwise, the label is 0 (negative pair). The contrastive loss, defined in Eq.~\textcolor{cyan}{(}\ref{eq:siameseloss}\textcolor{cyan}{)}, quantifies the similarity between the outputs of the two branches. Additionally, these new labels introduce a supplementary supervised learning component, further fine-tuning both tasks.

In summary, Combined-SSL integrates pretext tasks, downstream tasks, and Contrastive Learning, creating a unified framework that enhances feature distinction for both tasks. It roots in the chain rule, and seamlessly optimizes parameters across integrated tasks.

\subsection{Combined Batch- and Subject-level (CBS) Loss} \label{sec:3.CBSloss}

The SSL-V3 model employs batch- and subject-level loss functions (See Eq.~\textcolor{cyan}{(}\ref{eq:totaloss}\textcolor{cyan}{)}). The batch-level loss comprises Focal Loss and Contrastive loss. The subject-level loss is a Binary Cross Entropy (BCE) loss. Although this paper focuses on binary classification, the method can be extended to multi-class classification.

\vspace{-0.5cm}
\begin{flalign} 
\begin{split}\label{eq:totaloss}
L_{CBS} &= \text{Batch-level Loss} \\
&+ I*0.5*\text{Subject-level Loss} \\
&= \text{FL} + 0.5*\text{CL} + I*0.5*\text{BCE} 
\end{split} \\
I &= \begin{cases} 1 & \text{if last batch}\\ 0 & \text{otherwise} \end{cases} 
\end{flalign}
where $FL$ denotes Focal Loss, while $CL$ represents Contrastive loss. 
$I$ is an indicator that signifies whether it meets the last batch.

To be specific, If the dataset has inter- and intra-class imbalanced issues~\cite{S1_1-MC-ViViT_2023}, Focal Loss solves inter-class imbalanced issues, while Contrastive loss (See Eq.~\textcolor{cyan}{(}\ref{eq:siameseloss}\textcolor{cyan}{)}) addresses intra-class imbalanced ones.

\vspace{-0.5cm}
\begin{equation} \label{eq:siameseloss}
\begin{split}
&CL(\theta_{\text{CLS}},(M,\text{CLS}_{1}, \text{CLS}_{2})) = \frac{1}{2n}\sum^{n}_{i=1}M*D^{2}_{\theta_{\text{CLS}}} \\
&+ (1-M)\max(m-D_{\theta_{\text{CLS}}},0)^{2} \\
&D_{\theta_{\text{CLS}}}(\text{CLS}_{1}, \text{CLS}_{2}) = ||\text{CLS}_{1}-\text{CLS}_{2}||_{2} \\
&= (\sum^{P}_{j=1}(\text{CLS}^{j}_{1}-\text{CLS}^{j}_{2})^{2})^{1/2}
\end{split}
\end{equation}
where $\theta_{\text{CLS}}$ is the classification's parameters, $M$ indicates the matching label: specifically, $M=1$ if the label of $\mathbf{X}_{1}$ equals that of $\mathbf{X}_{2}$, otherwise $M=0$. 
Here, $n$ is the mini-batch size, and $D_{\theta_{CLS}}$ is the Euclidean distance of the two representation vectors. 
$m$ represents the threshold (m=2), while $P$ signifies feature dimension.

In SSL-V3, Contrastive loss measures the discrepancy between the two branches. Contrastive loss leverages the contrastive structure and examines the relationship between samples within the current batch as well. 
In addition, the Contrastive loss is characterized by less complex factors and a simpler format.
The BCE loss serves as a subject-level loss, computes the prediction loss subject by subject, and accumulates them together.

\section{Experiments} \label{sec:4}

In this section, we first introduce the datasets, evaluation metrics, and implementation details. Then, we conduct the experiments and ablation study, and evaluate the results.

\vspace{-0.3cm}
\subsection{Dataset and Processing} \label{sec:4.1}

\noindent \textbf{I-CONECT dataset}
The Internet-Based Conversational Engagement Clinical Trial (I-CONECT) contains a video dataset for the detection of Mild Cognitive Impairment (MCI) detection task\footnotemark{}\footnotetext{Website: \url{https://www.i-conect.org/}.}~\cite{S1_1-MC-ViViT_2023, S1_16-I-CONECT-2_2022}. 
Each sample is a 30-minute theme-oriented interview video between the interviewee (subject) and the interviewer. 
I-CONECT convened 186 subjects older than 75, including 100 with MCI and 86 with Normal Cognition (NC) (ClinialTrials.gov \#: NCT02871921).

However, the video quality varies significantly from subject to subject as they were recorded with different cameras, recording speeds and resolutions, and indoor lighting conditions. 
High-resolution videos result in more robust predictions than blurred ones~\cite{S1_1-MC-ViViT_2023}, which motivates us to study whether referring to video quality benefits prediction. 
Hence, I-CONECT is appropriate for testing the proposed SSL-V3 framework.

This paper focuses on 4 themes: Crafts Hobbies, Day Time TV Shows, Movie Genres, and School Subjects. Table~\ref{tab:theme_intro} presents the data distribution.

\begin{table} [ht]
\vspace{-0.3cm}
\centering
\caption{The details of researched themes. Subject Number is the total number of videos in the corresponding theme. Male/Female and MCI/NC show the gender distribution and category distribution. Frame Number is the total image number of each theme after converting videos into frames.}
\vspace{-0.2cm}
\begin{tabular}{lcccc}
\toprule
& \multirow{2}{1.5cm}{Crafts Hobbies} & \multirow{2}{2.cm}{Day Time TV Shows} & \multirow{2}{1.5cm}{Movie Genres} & \multirow{2}{1.5cm}{School Subjects} \\
& & & & \\ 
\midrule
Subject Number & 32 & 41 & 35 & 39 \\ 
Male/Female & 11/21 & 11/30 & 10/25 & 11/28 \\
MCI/NC & 20/12 & 20/21 & 21/14 & 22/17 \\
Frame Number & 691872 & 859872 & 770656 & 797968 \\
\bottomrule
\end{tabular}

\label{tab:theme_intro} \vspace{-0.3cm}
\end{table}  

\begin{figure}[ht]
\centering
\vspace{-0.4cm}
\subfloat[]{{\includegraphics[width=4cm]{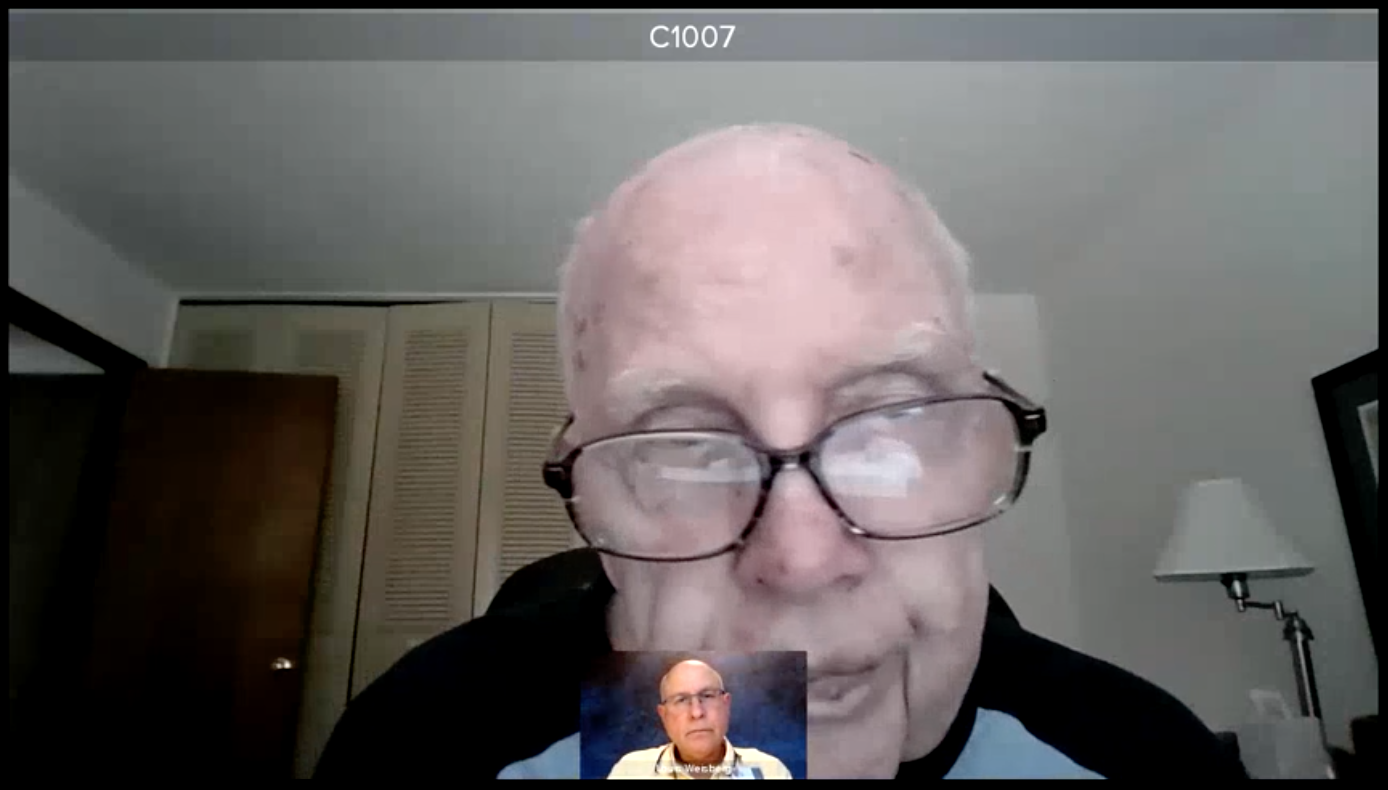} \label{fig:frame-l}}}
\subfloat[]{{\includegraphics[width=4cm]{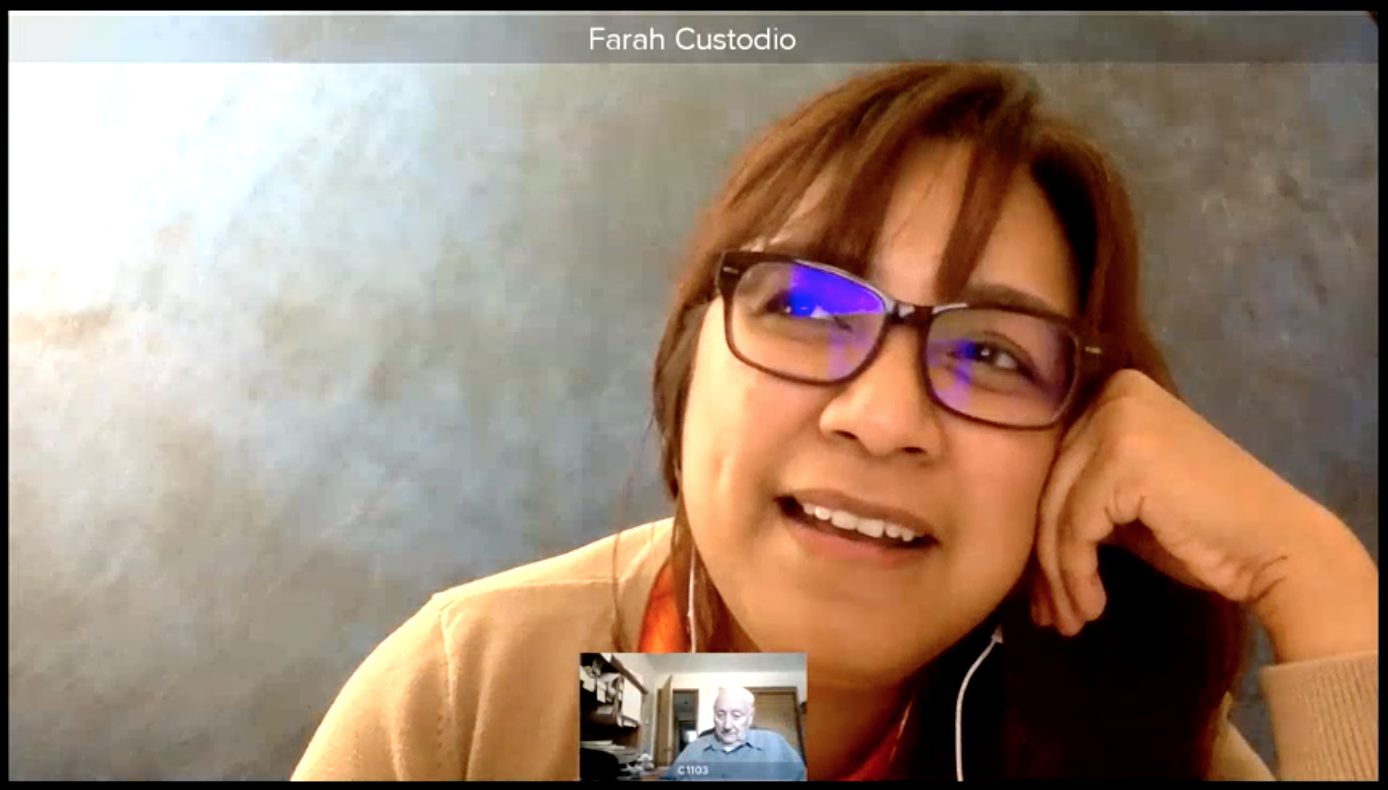}\label{fig:frame-s}}}
\caption{Two sample frames from the I-CONECT dataset. In (a), the window of the interviewee is bigger than that of the interviewer because the interviewer was speaking. Conversely, in (b), the interviewer was talking so that her window was bigger than the interviewee's.}
\label{fig:video-sample}
\vspace{-0.3cm}
\end{figure}

Typically, in each video, the subject exhibits normal cognition at the start and the end, introducing many noisy features that reduce accuracy. To avoid interfering, we discard the first 3 minutes and the last 2.5 minutes of each video. 
Then, Fig.~\ref{fig:video-sample} states that every video has a complex background and interviewer's face. 
To focus on the subject's facial expression, we employed EasyOCR+RetineFace~\cite{S4_1_1-retinaface_2020} to crop the interviewees' facial images and exclude all other information.

Simultaneously, inspired by~\cite{S4_1_3-FER-GCN_2021,S4_1_4-TSM_2021}, this study uses sequence-based approaches, taking consecutive frames as the input and setting the frame number as 16.
To fully leverage the videos, we use K-fold Cross Validation. 
Table~\ref{tab:exp_one_theme} specifies the fold number for each theme. Additionally, the videos in each fold belong to different subjects.

\noindent \textbf{Hockey Fight Detection dataset}

Fighting involves high-speed body motion, resulting in blurred and low-quality video, which can lead to mediocre predictions. To validate the generalization of SSL-V3, we also use the Hockey Fight Detection (HF) dataset~\cite{S4_1_6-HF_2011} as a secondary dataset to determine if SSL-V3 can achieve reasonable results.

\begin{figure}[ht]
\centering
\vspace{-0.5cm}
\subfloat[]{{\includegraphics[width=4.3cm,height=3cm]{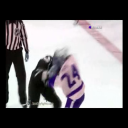}} \label{fig:fight}}
\subfloat[]{{\includegraphics[width=4.3cm,height=3cm]{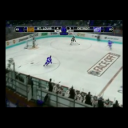}}\label{fig:nofight}}
\vspace{-0.2cm}
\caption{Two samples from HF dataset. (a) is the scenario of fighting action. (b) presents the normal non-fight situation.}
\label{fig:video-sample2}
\vspace{-0.1cm}
\end{figure}

The HF dataset assesses fight and violence in the hockey games of the National Hockey League. It contains 1000 video clips with a resolution of 360 $\times$ 288 pixels. Each video lasts between 1 and 2 seconds with a frame rate of 25 FPS. The HF dataset contains two categories: 500 fights and 500 non-fights. Fig.~\ref{fig:video-sample2} shows examples of both fight and non-fight clips.

Similar to the I-CONECT dataset, we also use sets of adjacent 16 frames as input. We then randomly divide the dataset into training, validation, and test sets in an 8:1:1 ratio, and report the mean and standard deviation (STD) values from five randomized experiments.

\subsection{Evaluation Metrics} \label{sec:4.2}

This work employs five typical metrics - Prediction accuracy, F1 score, AUC (Area Under the Receiver Operating Characteristic Curve), Sensitivity, and Specificity - to evaluate classification performance. 

\subsection{Implementation Details} \label{sec:4.3}

We split the input clips into several low-resolution cubic patches. Then, we augment the input cubes in various ways (random horizontal and vertical flip, random rotation, center crop, and color space augmentations) for two branches. Next, SSL-V3 shuffles and permutes the clips within the input mini-batch to generate the second input. Two branches accept the two inputs individually. Here, CBS Loss supervises the training process. The batch size is 80 for the I-CONECT dataset and 40 for the HF dataset, with an input frame length of 16. The initialized learning rate is 1e-8. We used the Adam optimizer and Cyclic scheduler with the triangular2 mode. The epoch number is 50. We implemented the model using PyTorch 1.12.0+cu116 and conducted experiments on an NVIDIA GTX 3090 GPU.

\subsection{MCI Detection} \label{sec:4.4}

MCI is an international public health issue. Worldwide, up to 15.56\% of community dwellers aged over 50 years are affected by MCI~\cite{S1_2-MCIrateworldwide_2022}. Furthermore, 21.2\% older adults in nursing homes have MCI~\cite{S1_3-MCI_nursehome_2023}. Efficient early detection of MCI is urgently needed. Hence, we apply SSL-V3 to this healthcare topic to test if SSL-V3 can efficiently detect MCI from NC in each theme.

\begin{table}[ht]
\vspace{-0.3cm}
\caption{The prediction accuracy of detecting MCI on 4 themes using K-fold evaluation.}
\vspace{-0.2cm}
\centering
\begin{tabular}{lcccc}
\hline
\multirow{2}{1.5cm}{Test Theme} & \multirow{2}{1.5cm}{\centering Crafts Hobbies} & \multirow{2}{2.cm}{\centering Day Time TV Shows} & \multirow{2}{1.5cm}{\centering Movie Genres} & \multirow{2}{1.5cm}{\centering School Subjects} \\
& & & & \\ 
\hline
Fold Num & 11 & 14 & 11 & 13 \\ 
\multirow{2}{1.5cm}{Accuracy} & 30/32 & 37/41 & 31/35 & 37/39 \\
\cmidrule{2-5}
& 93.75\% & 90.24\% & 88.57\% & 94.87\% \\
F1 & 95.00\% & 90.00\% & 91.30\% & 95.65\% \\
AUC & 93.33\% & 90.24\% & 88.90\% & 94.12\% \\
MCI/NC & 20/12 & 20/21 & 21/14 & 22/17 \\
Sensitivity/ & 95\%/ & 90\%/ & 95.45\%/ & 100\%/ \\
Specificity  & 91.67\% & 90.48\% & 82.35\% & 88.24\% \\
\hline
\end{tabular}
\vspace{-0.3cm}
\label{tab:exp_one_theme}
\end{table} 

Table~\ref{tab:exp_one_theme} shows the specific results. All accuracies from SSL-V3 are over 88\%. For example, on the theme of School Subjects, SSL-V3 successfully predicts 37 out of 39 cases, achieving the highest accuracy of 94.87\%. It also reaches 93.75\% accuracy on the Crafts Hobbies, which is close to 94\%. Additionally, the strong results of the rest metrics also solidly support that SSL-V3 has robust performance and can reliably and balancedly predict positive and negative samples with the help of video quality score. Table~\ref{tab:otherWork} also indicates that SSL-V3 outperforms the other models and referring to VQA significantly benefits the prediction.

\begin{table}[ht]
\vspace{-0.4cm}
\caption{Compare the performance of SSL-V3 with other models on the I-CONECT dataset. No VQA means to classify the videos without referring to the video quality, while VQA represents using VQA to classify videos.}
\vspace{-0.2cm}
\centering
\begin{tabular}{lccc}
\hline
 & Data Modality & Accuracy & F1 \\
\hline
Chen~\etal~\cite{S5_1-Conv-2020}   & Text & 79.15\% & - \\ 
Liu~\etal~\cite{S5_2-Lang-MRI-2022} & Text, MRI & 87.00\% & 89.00\% \\
Alsuhaibani~\etal~\cite{S4_1_6-Muath_2024} & Video & 87.50\% & 89.00\% \\
MC-ViViT~\citep{S1_1-MC-ViViT_2023} & Video & 90.63\% & 93.03\% \\
SSL-V3 (No VQA) & Video & 87.80\% & 86.49\% \\
SSL-V3 (VQA) & Video & \textbf{94.87\%} & \textbf{95.65\%} \\
\hline
\end{tabular}
\vspace{-0.3cm}
\label{tab:otherWork}
\end{table} 


\subsection{Violence Detection} \label{sec:4.4.1}

Violence detection, as a part of human action recognition at a distance, is an important topic in surveillance. It is crucial for public safety. Therefore, we validate SSL-V3 on the HF dataset to detect violent actions. 

\begin{table}[ht]
\vspace{-0.5cm}
\caption{The prediction accuracy of detecting violence on HF dataset. The report format is mean $\pm$ STD\%}
\vspace{-0.3cm}
\centering
\begin{tabular}{ccccc}
\hline
Accuracy & F1 & AUC & Sensitivity & Specificity \\
\hline
98.6$\pm$0.8\% & 98.59$\pm$0.81\% & 98.6$\pm$0.8\% & 98.0$\pm$1.26\% & 99.2$\pm$0.99\% \\
\hline
\end{tabular}
\vspace{-0.5cm}
\label{tab:exp_hf}
\end{table} 

Table~\ref{tab:exp_hf} presents that SSL-V3 achieved promising results on all five metrics, all mean values are over 98\% with low standard deviations (STD), indicating the robustness of SSL-V3 in detecting positive and negative samples. Moreover, according to Table~\ref{tab:otherWork_hf}, our mean accuracy of 98.6\% with an STD of 0.008 is comparable to the top-1 accuracy achieved by SepConvLSTM-C~\cite{S5_12-SepConvLSTM-C_2021} (99.5\%). The joining of VQA devoted to the good result a lot. Therefore, SSL-V3 can efficiently detect fighting during hockey games.

\begin{table}[ht]
\vspace{-0.4cm}
\caption{Compare the performance of other models with SSL-V3 on the HF dataset.}
\vspace{-0.2cm}
\centering
\begin{tabular}{lll}
\hline
 & Accuracy & F1 \\
\hline
CNN+SVM~\citep{S5_5-CNN+SVM_2021} & 93.7\% & - \\
2D CNN~\citep{S5_6-HF-2DCNN_2018} & 94.60$\pm$0.6\% & - \\ 
MobileNet~\citep{S5_7-MobileNet_2022} & 96.66\% & - \\
C3D+SVM~\citep{S5_8-C3D+SVM_2021} & 97.86\% & 97.87\% \\
\multirow{2}{4cm}{Computationally Intelligent VD~\citep{S5_9-VD_2022}} & \multirow{2}{1cm}{98\%} & \multirow{2}{1cm}{98.10\%} \\
& & \\
Spatial Encoder~\citep{S5_10-Spatial-Encoder_2018} & 98.1$\pm$0.58\% & - \\
3D CNN~\citep{S5_11-3DCNN_2019} & 98.3$\pm$0.81\% & - \\
SepConvLSTM-C~\citep{S5_12-SepConvLSTM-C_2021} & 99.5\% & - \\
SSL-V3 (No VQA) & 95.0$\pm$1.41\% & 94.98$\pm$1.36\% \\
SSL-V3 (VQA) & 98.6$\pm$0.8\% & 98.59$\pm$0.81\% \\
\hline
\end{tabular}
\vspace{-0.3cm}
\label{tab:otherWork_hf}
\end{table}

\subsection{Ablation Study} \label{sec:4.5}

The ablation study is conducted on the I-CONECT and HF datasets.

\subsubsection{Study the essential of VQA task} \label{sec:4.5.1}

This study investigates the impact of the VQA task. Testing the model without the intervention of VQA involves using neither SSR nor VSR, the first row of Table~\ref{tab:tune_HEAD}. 

\begin{table}[ht]
\vspace{-0.3cm}
\caption{The prediction accuracy on 4 I-CONECT themes and HF dataset with different regressors. SSR and VSR denotes Sequence Score Regressor and Video Score Regressor respectively.}
\vspace{-0.2cm}
\centering
\begin{tabular}{ccccccc}
\hline
\multirow{2}{0.5cm}{SSR} & \multirow{2}{0.5cm}{VSR} & \multirow{2}{1.cm}{Crafts Hobbies} & \multirow{2}{2.cm}{Day Time TV Shows} & \multirow{2}{1.cm}{Movie Genres} & \multirow{2}{1.1cm}{School Subjects} & \multirow{2}{1.1cm}{Hockey Fight} \\
& & & & & & \\ 
\hline
\XSolidBrush & \XSolidBrush & 84.38\% & 87.80\% & 80.00\% & 82.05\% & (95$\pm$1.41)\% \\
\Checkmark   &  & 87.50\% & 78.05\% & 88.57\% & 84.62\% & (95.6$\pm$1.36)\% \\
 & \Checkmark & 87.50\% & 87.80\% & 82.86\% & 76.92\% & (95.6$\pm$2.5)\% \\
\Checkmark   & \Checkmark & \textbf{93.75\%} & \textbf{90.24\%} & \textbf{88.57\%} & \textbf{94.87\%} & \textbf{(98.6$\pm$0.8)\%} \\
\hline
\end{tabular}
\vspace{-0.3cm}
\label{tab:tune_HEAD}
\end{table} 

Table~\ref{tab:tune_HEAD} demonstrates that, across all themes, VQA head contributes to significant improvement in the performance of SSL-V3. For instance, the VQA head provides more than 8.5\% improvement in accuracy for SSL-V3 in the Crafts Hobbies and Movie Genres. 
The increase is even 12.82\% in the School Subjects. Therefore, referring to VQA significantly benefits the model's prediction accuracy, underscoring the necessity of considering video quality in video classification. 
The results from the HF dataset also support this conclusion (Table~\ref{tab:tune_HF_all}). 

At the same time, Table~\ref{tab:tune_HEAD} supports that VQA auxiliary improves the learned representation and downstream performance instead of causing harmful over-fitting to augmentation artifacts. 
Importantly, in all experiments, no data augmentation is applied at inference time, yet the model consistently achieves strong performance.
Since the VQA head operates on clean inputs at inference, if it had relied on augmentation-specific artifacts learned during training, its outputs and the resulting classification would be expected to degrade under this distribution shift; however, this behavior is not observed.
Hence, augmentation artifacts are incorporated as controlled, task-aligned degradation signals (not contradictory supervision), and both the method design and ablation Table~\ref{tab:tune_HEAD}, together with the strong clean-inference performance, support that they regularize - rather than shortcut - the learned representation

\subsubsection{Study the best structure of VQA head} \label{sec:4.5.2}

This study investigates the necessity of VQA head's two modules, SSR and VSR. Tang~\etal~\cite{S1_11-VGG-16_2020} and Kim~\etal~\cite{S1_12-DeepVR-VQA_2020} tuned the parameters of each module respectively. 

In the case of using VSR alone, SSR is replaced with a 1D average pooling layer, which operates on the extracted features of each sequence. Conversely, when using SSR alone, the VSR module is replaced with a single FC layer that directly regresses the output of SSR to the overall clip quality score.

Table~\ref{tab:tune_HEAD} indicates that using the VQA head results in better performance than using either SSR or VSR alone. For example, the accuracy of the VQA head is consistently at least 2.44\% higher than that of VSR alone. While the VQA head and SSR alone achieve comparable accuracy in Movie Genres, the VQA head consistently outperforms SSR alone by at least 6.25\% across other themes.

\begin{figure}[ht]
\vspace{-0.4cm}
\begin{center}
\includegraphics[width=7.7cm,height=5.cm]{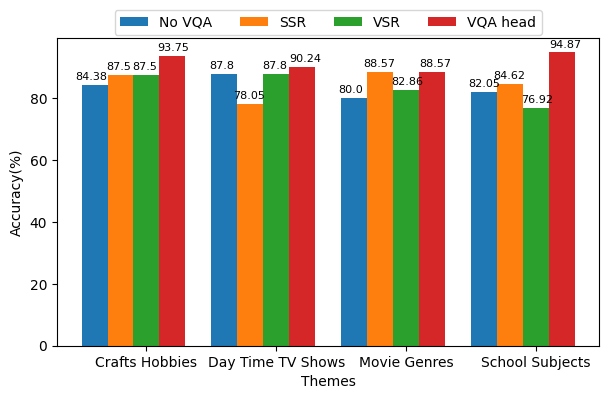}
\end{center}
\vspace{-0.6cm}
\caption{The prediction accuracy on 4 themes (I-CONECT dataset) with different regressors.}
\label{fig:Abl-Head}
\vspace{-0.1cm}
\end{figure}

Between the two modules, SSR alone performs better than VSR alone in Movie Genres and School Subjects and ties with VSR alone in Crafts Hobbies. Specifically, in School Subjects, adding the VSR module brings a 10.25\% accuracy increase to the SSL-V3, while adding SSR provides approximately an 18\% improvement in accuracy. This highlights the effectiveness of SSR in enhancing performance over VSR. Fig.~\ref{fig:Abl-Head} reflects this disparity. SSR learns raw Spatio-Temporal features, whereas replacing SSR with 1D average pooling removes specific characteristics from the features, hindering VSR's ability to make precise decisions. Hence, SSR alone outperforms VSR alone. We draw the same conclusion from HF dataset. The specific results in Tables~\ref{tab:tune_HEAD_ALL} and~\ref{tab:tune_HF_all} from Appendix~\ref{app:2} support this view as well.

Overall, the VQA head enhances SSL-V3's performance substantially compared to using either SSR or VSR alone. Within the VQA head, SSR is more effective than VSR.

\subsubsection{Study the effectiveness of contrastive structure} \label{sec:4.5.3}

This study inspects the essence of using the contrastive structure in Combined-SSL. Discarding the contrastive structure involves removing the Contrastive loss from the CBS Loss. 

\begin{table}[ht]
\vspace{-0.3cm}
\caption{The prediction accuracy on 4 I-CONECT themes and HF dataset with different loss functions. CL means Contrastive Learning.}
\vspace{-0.2cm}
\centering
\begin{tabular}{cccccccc}
\hline
\multirow{2}{0.5cm}{FL} & \multirow{2}{0.5cm}{CL Loss} & \multirow{2}{0.5cm}{BCE Loss} & \multirow{2}{1.cm}{Crafts Hobbies} & \multirow{2}{2.cm}{Day Time TV Shows} & \multirow{2}{1.cm}{Movie Genres} & \multirow{2}{1.1cm}{School Subjects} & \multirow{2}{1.1cm}{Hockey Fight} \\
& & & & & & & \\ 
\hline
\Checkmark &         &         & 81.25\% & 65.85\% & 54.29\% & 51.28\% & (92.8$\pm$1.6)\% \\
\Checkmark & \Checkmark &         & 84.38\% & 85.37\% & 71.43\% & 79.49\% & (94.4$\pm$1.96)\% \\
\Checkmark &         & \Checkmark & 81.25\% & 68.29\% & 71.43\% & 69.23\% & (94.2$\pm$1.47)\% \\
\Checkmark & \Checkmark & \Checkmark & \textbf{93.75\%} & \textbf{90.24\%} & 8\textbf{8.57\%} & \textbf{94.87\%} & \textbf{(98.6$\pm$0.8)\%} \\
\hline
\end{tabular}
\vspace{-0.5cm}
\label{tab:tune_LOSS}
\end{table} 

Table~\ref{tab:tune_LOSS} shows that the contrastive structure substantially enhances the performance of SSL-V3. Across all themes, using the contrastive structure enables SSL-V3 to be at least 12.5\% more accurate than when it is discarded. This accords to the pattern of the results from HF dataset. Contrastive Learning is indispensable for Combined-SSL.

\subsubsection{Study suitable loss function} \label{sec:4.5.4}

This study evaluates the effect of various loss functions to determine the best one for SSL-V3. FL is the base of CBS Loss, with Contrastive loss and BCE losses serving as supplementary components. Except for purely FL, we combine FL with Contrastive loss and BCE alternately to test the effects of the two combinations.

Table~\ref{tab:tune_LOSS} shows that CBS Loss outperforms the other combination on both datasets. For instance, in Movie Genres, CBS Loss achieves its lowest accuracy at 88.57\%, which is 34.28\% larger than the accuracy of FL alone and approximately 17.1\% higher than those of FL + BCE and FL + Contrastive loss, both at 71.43\%. This disparity indicates the significance of using Contrastive loss and Subject-level loss. 

\begin{figure}[ht]
\vspace{-0.2cm}
\begin{center}
\includegraphics[width=7.7cm,height=5.cm]{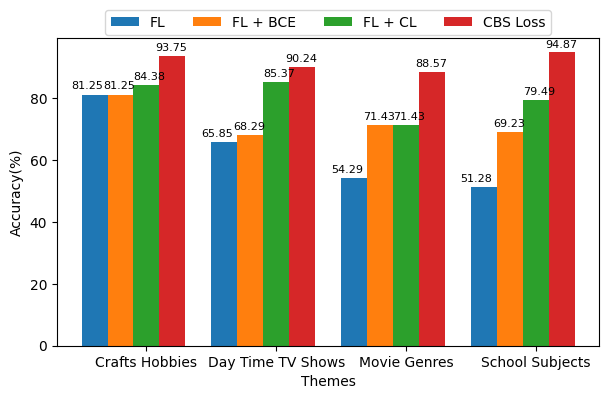}
\end{center}
\vspace{-0.6cm}
\caption{The prediction accuracy on 4 themes (I-CONECT dataset) with different loss functions. CL means Contrastive Learning.}
\label{fig:Abl-Loss}
\vspace{-0.1cm}
\end{figure}

Both FL + BCE and FL + Contrastive loss outperform FL alone across the overall themes (Fig.~\ref{fig:Abl-Loss}). 
With Contrastive loss included, FL + Contrastive loss achieved 3.13\% and 10.26\% higher accuracy on Crafts Hobbies and School Subjects, respectively, compared to using FL + BCE. 
Thus, Contrastive loss is important, also supported by the results of HF dataset, highlighting that the batch-level loss forms the foundation part. 
When BCE loss, which operates at the subject level, is incorporated, SSL-V3 demonstrates improved performance. 
Without Contrastive loss, empowering the model and enhancing predictions at the subject level becomes challenging, largely due to unresolved intra-class imbalance issues. 
Contrastive loss effectively addresses intra-class imbalances within the mini-batch context. 
Furthermore, discarding the Contrastive loss is equivalent to removing the contrastive structure, which substantially decreases the performance of SSL-V3, demonstrating the importance of Contrastive loss. 
All results in Tables~\ref{tab:tune_LOSS_ALL} and~\ref{tab:tune_HF_all} from Appendix~\ref{app:2} strongly support this conclusion.

In summary, the results of the ablation study support selecting CBS Loss.

\subsubsection{Compare the impact of CBS Loss to that of VQA head} \label{sec:4.5.5}

Table~\ref{tab:tune_LOSS} implies that with the VQA head and partial CBS Loss, SSL-V3 struggles to achieve over 80\% accuracy in all themes except Crafts Hobbies.

Table~\ref{tab:tune_HEAD} illustrates that with CBS Loss and a partial VQA head, SSL-V3 achieves over 80\% accuracy effortlessly. 
Despite this, VSR alone achieves 76.92\% on School Subjects, which ranks as the fourth highest value among FL alone, FL+BCE, and FL+Contrastive Learning in Table~\ref{tab:tune_LOSS}.

Hence, comparing the two tables indicates that CBS Loss has a greater impact on SSL-V3 than the VQA head does. 
In essence, a robust loss function is more crucial than a high-performing classifier or regressor head.

\section{Discussion} \label{sec:5}

SSL-V3 is designed for two tasks: VQA and classification, making it a potential multi-task model. Both tasks train the same ViViT together, driving it to learn video content from diverse perspectives. Moreover, the collaboration of each module also leads to SSL-V3's robust performance in MCI detection and violence detection.

Specifically, \textbf{Combined-SSL} optimizes VQA and classification reciprocally, effectively addressing the label shortage in the VQA task. In addition, the contrastive structure helps to tune the VQA task comprehensively and emphasizes making features from the same group denser and those from opposite classes more separable. \textbf{VQA head} highlights the effect of important features by the SSR module and fuses the temporal effect across different channels stereoscopically. 

\begin{figure}[ht]
\begin{center}
\subfloat[]{{\includegraphics[width=4.4cm]{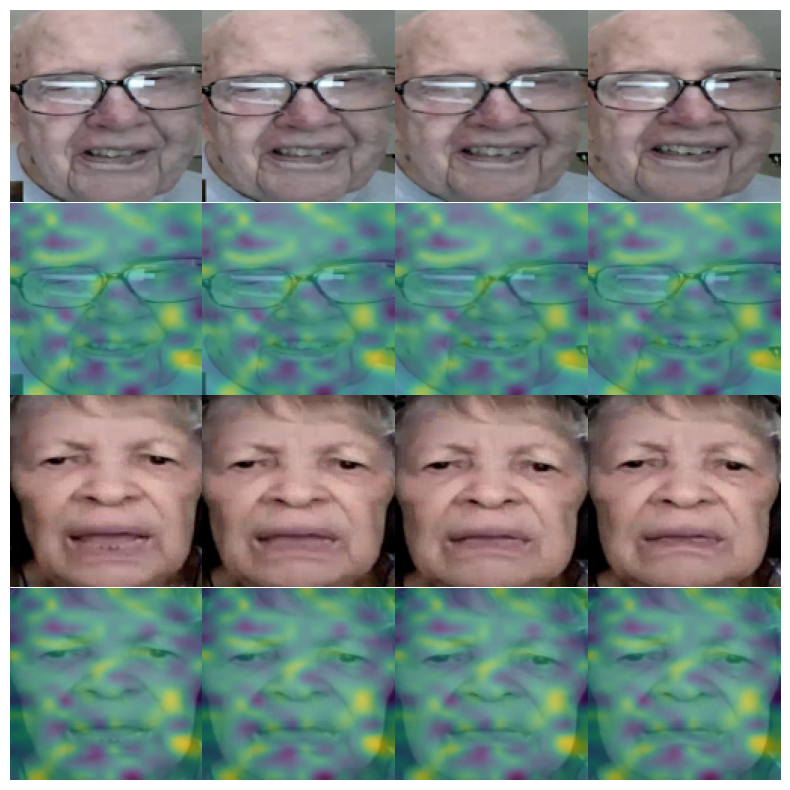} \label{fig:hm_iconect}}}
\subfloat[]{{\includegraphics[width=4.4cm]{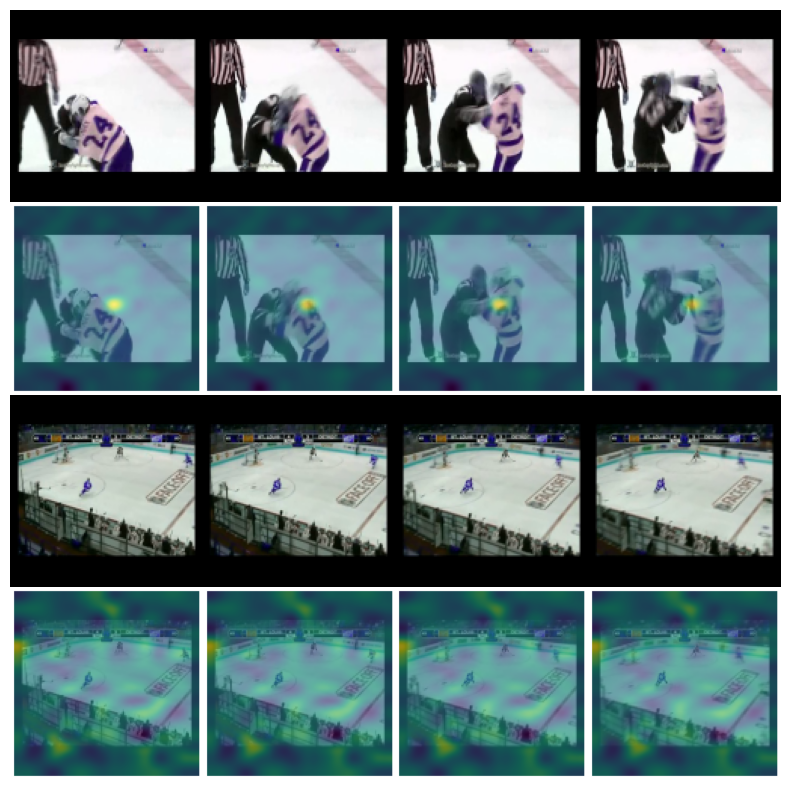} \label{fig:hm_hf}}}
\end{center}
\vspace{-0.4cm}
\caption{The extracted spatio-temporal feature maps on 16 consecutive frames from two datasets. SSL-V3 focuses more on the bright areas and less on the dark ones. Given that there are $n_{t}$ tubelets, where $n_{t}$=$\frac{T}{t}$=$\frac{16}{4}$=$4$, there are four feature maps. To simplify the visualization, we pick the first frame of each tubelet to represent it. Fig.~\ref{fig:hm_iconect} is from the I-CONECT dataset, while Fig.~\ref{fig:hm_hf} is from HF one.}
\label{fig:VVT-HM}
\vspace{-0.3cm}
\end{figure}

\textbf{ViViT}, with the shared weight mechanism, extracts a diverse and rich set of Spatio-Temporal features to meet the demands of VQA and classification. Fig.~\ref{fig:VVT-HM} shows that ViViT pays particular attention to muscle motion in the cheek and brow bone regions within the I-CONECT dataset (Fig.~\ref{fig:hm_iconect}). ViViT also focuses on muscle motions in other areas like the forehead, nose, eyelids, and lips. Additionally, in the HF dataset, Fig.~\ref{fig:hm_hf} suggests that ViViT highlights body motion in the fighting video. In contrast, in non-fighting videos, SSL-V3 tends to capture the movement of single objects and the environment of the court. 

The \textbf{CBS Loss} optimizes SSL-V3 at both Batch and Subject levels. Notably, using the contrastive structure, CBS loss solves the intra-class imbalance issue at the mini-batch level. Additionally, it requires no memory bank, making it space- and computation-efficient.

In summary, there are five key findings or strengths. 
\begin{itemize}
\item Referring to video quality scores benefits the neural network in classification over existing models. 

\item Noval Combined-SSL alleviates the pressure of label scarcity in VQA. 

\item Contrastive Learning improves model performance and addresses class imbalance issues. 

\item SSL-V3 has strong generalization across domains (healthcare and surveillance).

\item SSL-V3 is scalable. ViViT's property on sequential video processing enables the SSL-V3 to classify longer video and execute real-time constraint on each clip.
\end{itemize}

\section{Conclusion and Future Work} \label{sec:6}

This paper developed SSL-V3 based on the proposed Combined-SSL to perform video classification by referring to the video quality score. Combined-SSL uses video quality score as a factor to tune the classification task and applies a contrastive structure to improve performance. SSL-V3 also impresses people by practical applicability. The good experimental results from the I-CONECT and HF datasets verify the effectiveness of SSL-V3 on real-world noisy video streams in healthcare and surveillance. Considering video quality is the right move while doing classification. 

Simultaneously, we would like to highlight that different from Pretext Task and Contrastive Learning, Combined-SSL drops the fine-tuning part. Moreover, the designing logic behind Combined-SSL offers a new way to address the label shortage issue, which is combining the target task with one or even more tasks with sufficient labels.

However, this study has limitations. 
\begin{itemize} 
\item \textbf{SSL-V3 is not a fully multi-task model}

Due to missing labels, it is difficult to evaluate the computed video quality score using metrics like Spearman's Rand-order Correlation Coefficient, Pearson Linear Correlation Coefficient, and Root Mean Squared Error. Hence, it is hard to claim that SSL-V3 is a fully multi-task model. 

\item \textbf{Referring to the video quality score is not panacea}

The prediction of all perfect-quality video is surely better than that of referring to the video quality score.

\item \textbf{Potential overfitting}

The SSL-V3 may suffer from overfitting issue while processing datasets with small subject number and short video clips.
\end{itemize}

Hence, future work will focus on experimenting with SSL-V3 on public video datasets with both category labels and video quality scores to verify that SSL-V3 is a generalized SSL multi-task model. Then, the importance of video quality suggests that even a simple neural network can achieve good accuracy with high-quality video. Another research direction is to perform video denoising using a generative model before classification. This way helps skip the video quality part and generate high-quality video, promising good and objective performance for sure.

\backmatter

\bmhead{Acknowledgments}

The collection of the I-CONECT dataset was partly supported by the National Institute of Health (NIH) funding: R01AG051628 and R01AG056102.

\section*{Declarations}

\textbf{Funding} Not applicable

\textbf{Conflict of interest} The authors have no relevant financial or non-financial interests to disclose.

\textbf{Ethics approval} Not applicable

\textbf{Consent for publication} Not applicable

\textbf{Data availability} I-CONECT dataset will be made available on request through \url{https://www.i-conect.org/}.

Hockey Fight Detection Dataset is a public dataset 

(\url{https://paperswithcode.com/dataset/hockey-fight-detection-dataset}).

\textbf{Materials availability} Not applicable



\begin{appendices}
\section{Full Results of Ablation Study} \label{app:2}

\begin{table}[ht]
\vspace{-0.3cm}
\caption{Specific results on 4 themes with different loss functions. CL means Contrastive Learning.}
\vspace{-0.4cm}
\centering
\begin{tabular}{@{\extracolsep\fill}lcccccc}
\hline
\multirow{3}{1.1cm}{Test Themes} & \multicolumn{3}{c}{Crafts Hobbies} & \multicolumn{3}{c}{Day Time TV Shows} \\
\cmidrule{2-7}
& \multirow{2}{1.3cm}{\centering FL} & \multirow{2}{1.3cm}{\centering FL + BCE} & \multirow{2}{1.1cm}{\centering FL + CL} & \multirow{2}{1.3cm}{\centering FL} & \multirow{2}{1.3cm}{\centering FL + BCE} & \multirow{2}{1.1cm}{\centering FL + CL} \\
 & & & & & & \\
\hline
\multirow{2}{1.5cm}{Accuracy} & 26/32 & 26/32 & 27/32 & 27/41 & 28/41 & 28/41 \\
\cmidrule{2-7}
& 81.25\% & 81.25\% & \textbf{84.38\%} & 65.85\% & 68.29\% & \textbf{85.37\%} \\
F1 & 82.35\% & 84.21\% & \textbf{85.71\%} & 50.00\% & 68.29\% & \textbf{85.00\%} \\
AUC & 85.00\% & 81.67\% & \textbf{87.50\%} & 65.12\% & 68.33\% & \textbf{85.36\%} \\
MCI/NC & \multicolumn{3}{c}{20/12} & \multicolumn{3}{c}{20/21} \\
Sensitivity/ & 70.00\%/ & \textbf{80.00\%}/ & 75.00\%/ & 35.00\%/ & 70.00\%/ & \textbf{85.00\%}/ \\
Specificity & 100.00\% & \textbf{83.33\%} & 100.00\% & 95.24\% & 66.67\% & \textbf{85.71\%} \\
\hline

\multirow{3}{1.1cm}{Test Themes} & \multicolumn{3}{c}{Movie Genres} & \multicolumn{3}{c}{School Subjects} \\
\cmidrule{2-7}
& \multirow{2}{1.3cm}{\centering FL} & \multirow{2}{1.3cm}{\centering FL + BCE} & \multirow{2}{1.1cm}{\centering FL + CL} & \multirow{2}{1.3cm}{\centering FL} & \multirow{2}{1.3cm}{\centering FL + BCE} & \multirow{2}{1.1cm}{\centering FL + CL} \\
 & & & & & & \\
\hline
\multirow{2}{1.5cm}{Accuracy} & 19/35 & 25/35 & 25/35 & 20/39 & 27/39 & 31/39 \\
\cmidrule{2-7}
& 54.29\% & \textbf{71.43\%} & \textbf{71.43\%} & 51.28\% & 69.23\% & \textbf{79.49\%} \\
F1 & 46.66\% & \textbf{77.27\%} & \textbf{77.27\%} & 38.71\% & 75.00\% & \textbf{81.82\%} \\
AUC & 59.52\% & \textbf{69.05\%} & \textbf{69.05\%} & 53.57\% & 68.25\% & \textbf{79.37\%} \\
MCI/NC & \multicolumn{3}{c}{21/14} & \multicolumn{3}{c}{22/17} \\
Sensitivity/ & 33.33\%/ & \textbf{80.95\%}/ & \textbf{80.95\%}/ & 27.27\%/ & 81.82\%/ & \textbf{81.82\%}/ \\
Specificity & 85.71\% & \textbf{57.14\%} & \textbf{57.14\%} & 82.35\% & 52.94\% & \textbf{76.47\%} \\
\hline
\end{tabular}
\vspace{-0.4cm}
\label{tab:tune_LOSS_ALL}
\end{table} 

Table~\ref{tab:tune_LOSS_ALL} states the full results of tuning loss functions. FL + Contrastive loss achieves better Accuracy, F1, and Specificity than FL + BCE in the Crafts Hobbies and School Subjects themes. In the remaining two themes, FL + Contrastive loss and FL + BCE are tied. Overall, FL + Contrastive loss surpasses FL + BCE, indicating that addressing class imbalance issues takes priority over considering subject-level loss. Contrastive loss is more critical.

\begin{table}[ht]
\vspace{-0.3cm}
\caption{Specific results on 4 themes with different regression head.}
\vspace{-0.4cm}
\centering
\begin{tabular}{@{\extracolsep\fill}lcccccc}
\hline
\multirow{3}{1.1cm}{Test Themes} & \multicolumn{3}{c}{Crafts Hobbies} & \multicolumn{3}{c}{Day Time TV Shows} \\
\cmidrule{2-7}
& \multirow{2}{1.3cm}{\centering No VQA} & \multirow{2}{1.3cm}{\centering VSR} & \multirow{2}{1.3cm}{\centering SSR} & \multirow{2}{1.3cm}{\centering No VQA} & \multirow{2}{1.3cm}{\centering VSR} & \multirow{2}{1.3cm}{\centering SSR} \\
& & & & & & \\
\hline
\multirow{2}{1.5cm}{Accuracy} & 27/32 & 28/32 & 28/32 & 36/41 & 36/41 & 32/41 \\
\cmidrule{2-7}
& 84.38\% & \textbf{87.50\%} & \textbf{87.50\%} & \textbf{87.80\%} & \textbf{87.80\%} & 78.05\% \\
F1 & 87.18\% & \textbf{90.00\%} & \textbf{90.00\%} & 86.49\% & \textbf{87.18\%} & 78.05\% \\
AUC & \textbf{86.67\%} & \textbf{86.67\%} & \textbf{86.67\%} & \textbf{87.62\%} & \textbf{87.74\%} & 78.10\% \\
MCI/NC & \multicolumn{3}{c}{20/12} & \multicolumn{3}{c}{20/21} \\
Sensitivity & 85.00\% & \textbf{90.00\%} & \textbf{90.00\%} & 80.00\% & \textbf{85.00\%} & 80.00\% \\
Specificity & 83.33\% & \textbf{83.33\%} & \textbf{83.33\%} & 95.24\% & \textbf{90.48\%} & 76.19\% \\
\hline
\multirow{3}{1.1cm}{Test Themes} & \multicolumn{3}{c}{Movie Genres} & \multicolumn{3}{c}{School Subjects} \\
\cmidrule{2-7}
& \multirow{2}{1.3cm}{\centering No VQA} & \multirow{2}{1.3cm}{\centering VSR} & \multirow{2}{1.3cm}{\centering SSR} & \multirow{2}{1.3cm}{\centering No VQA} & \multirow{2}{1.3cm}{\centering VSR} & \multirow{2}{1.3cm}{\centering SSR} \\
& & & & & & \\
\hline
\multirow{2}{1.5cm}{Accuracy} & 28/35 & 29/35 & 31/35 & 32/39 & 30/39 & 33/39 \\
\cmidrule{2-7}
& 80.00\% & 82.86\% & \textbf{88.57\%} & 82.05\% & 76.92\% & \textbf{84.62\%} \\
F1 & 82.05\% & 86.96\% & \textbf{90.48\%} & 82.92\% & 75.68\% & \textbf{85.72\%} \\
AUC & 80.95\% & 79.76\% & \textbf{88.10\%} & 84.92\% & 78.17\% & \textbf{84.92\%} \\
MCI/NC & \multicolumn{3}{c}{21/14} & \multicolumn{3}{c}{22/17} \\
Sensitivity & 76.19\% & 95.24\% & \textbf{90.48\%} & 77.27\% & 63.64\% & \textbf{81.82\%} \\
Specificity & 85.71\% & 64.29\% & \textbf{85.71\%} & 88.24\% & 94.12\% & \textbf{88.24\%} \\
\hline
\end{tabular}
\vspace{-0.5cm}
\label{tab:tune_HEAD_ALL}
\end{table} 

Table~\ref{tab:tune_HEAD_ALL} contains all the metrics for tuning the regressor head. SSR and VSR reached better metric values than No VQA across the board. Focusing on SSR and VSR, in Crafts Hobbies and Day Time TV Shows, VSR alone performs better than or identical to SSR alone across all the metrics. However, in Movie Genres and School Subjects, SSR alone surpasses VSR alone in Accuracy and F1 score. They are tied in AUC. Furthermore, SSR alone achieves better Specificity in Movie Genres and higher Sensitivity in School Subjects than VSR alone. In summary, SSR alone leads to better performance than VSR alone, implying that SSR contributes more to SSL-V3.

\begin{table}[ht]
\vspace{-0.3cm}
\caption{Specific results on HF dataset with different loss functions and regressors. CL means Contrastive Learning.}
\label{tab:tune_HF_all}
\vspace{-0.2cm}
\centering
\begin{tabular}{@{\extracolsep\fill}lccc}
\hline
\multirow{3}{1.1cm}{Test Themes} & \multicolumn{3}{c}{Study Loss Function} \\
\cmidrule{2-4}
& \multirow{2}{1.3cm}{\centering FL} & \multirow{2}{1.3cm}{\centering FL + CL} & \multirow{2}{1.6cm}{\centering FL + BCE} \\
 & & & \\
\hline
Accuracy & (92.80$\pm$1.60)\% & \textbf{(94.40$\pm$1.96)\%} & (94.20$\pm$1.47)\% \\
F1 & (92.92$\pm$1.74)\% & \textbf{(94.38$\pm$1.96)\%} & (94.19$\pm$1.47)\% \\
AUC & (92.80$\pm$1.60)\% & \textbf{(94.40$\pm$1.96)\%} & (94.20$\pm$1.47)\% \\
Sensitivity/ & (90.80$\pm$3.71)\%/ & (94.00$\pm$4)\%/ & \textbf{(94.00$\pm$2.53)\%}/ \\
Specificity & (94.80$\pm$2.04)\% & (94.80$\pm$4.31)\% & \textbf{(94.40$\pm$2.93)\%} \\
\hline
\multirow{3}{1.1cm}{Test Themes} & \multicolumn{3}{c}{Study VQA head} \\
\cmidrule{2-4}
& \multirow{2}{1.3cm}{\centering No VQA} & \multirow{2}{1.3cm}{\centering SSR} & \multirow{2}{1.3cm}{\centering VSR} \\
 & & & \\
\hline
Accuracy & (95.00$\pm$1.41)\% & \textbf{(96.40$\pm$1.36)\%} & (95.60$\pm$2.50)\% \\
F1 & (94.98$\pm$1.36)\% & \textbf{(96.41$\pm$1.36)\%} & (95.56$\pm$2.53)\% \\
AUC & (94.60$\pm$1.36)\% & \textbf{(96.40$\pm$1.36)\%} & (95.60$\pm$2.50)\% \\
Sensitivity/ & (94.40$\pm$2.33)\%/ & \textbf{(96.80$\pm$1.60)\%}/ & (94.80$\pm$2.99)\%/ \\
Specificity & (95.60$\pm$3.20)\% & \textbf{(96.00$\pm$1.26)\%} & (96.40$\pm$2.65)\% \\
\hline
\end{tabular}
\vspace{-0.4cm}
\end{table}

Table~\ref{tab:tune_HF_all} contains all the results of ablation study from HF dataset. We can draw the same conclusion as Tables~\ref{tab:tune_LOSS_ALL} and~\ref{tab:tune_HEAD_ALL} do.

\section{What does last batch mean? How does per-subject aggregation?} \label{app:3}

Supposing there are 1000 batches in one epoch from batch 1 to batch 1000. The 1000th batch is the last batch in our case.

For a given batch, it contains clips from different subjects. 
To compute the subject-level loss, it requires to collect the prediction results of all clips. 
In the last batch of the current epoch (like 1000th batch), we finish collecting the results of all clips for each subject. 
Then we can start computing subject-level loss.

Supposing batch size = 4, $\hat{Y}_{b_{i}}=[s_{10}, s_{5}, s_{15}, s_{20}]$, $GT_{b_{i}}=[gt_{1}, gt_{2}, gt_{3}, gt_{4}]$, where $\hat{Y}_{b_{i}}$ is the predicted results of $i^{th}$ batch $b_{i}$, $s_{j}$ is each predicted subject index, $GT_{b_{i}}$ is the ground truth vector of $i^{th}$ batch $b_{i}$, $gt_{j}$ is each ground truth category ($i,j\in \mathbf{N}^{+}$).

If $s_{10}=gt_{1}$, we say the model predicts this clip to a correct category and mark it as 1. 
Otherwise, it is 0. We create a list $\text{acc\_num}$ to accumulately count the correct predicted number for each subject batch by batch. 
In the meanwhile, we use list $\text{cnt}$ to count all clip number of each subject.

In the last batch, we use this formula, $acc\_total = \text{round}(acc\_num/cnt)$, to calculate the subject-level accuracy $acc\_total$ for each subject. 
If the accuracy of certain subject is larger than 0.5, we round it to 1 (correct). Otherwise, it is 0 (wrong). 

Then, we use $acc\_total$ and overall ground truth result to compute binary cross entropy loss as the subject-level loss.

\section{Glossary} \label{app:1}

Table~\ref{tab:glssry} explains the detail of acronyms in this paper.

\begin{table}[ht]
\vspace{-0.3cm}
\caption{Glossary}
\vspace{-0.2cm}
\centering
\begin{tabular}{ll}
\hline
\multirow{2}{1.7cm}{Initials} & \multirow{2}{1.cm}{Vocabulary} \\
 & \\
\hline
AD & Alzheimer’s Disease \\
ADRD & Alzheimer’s Disease Related Dementia \\
BEEF & Bidirectional Effect-Based Fusion \\
\multirow{2}{1.5cm}{CBS Loss} & \multirow{2}{4.5cm}{Combined Batch- and Subject-level Loss} \\
 & \\
CLS & Classification \\
HF & Hockey Fight Detection \\
MC & Multi-branch Classifier \\
MCI & Mild Cognitive Impairment \\
MOS & Mean Option Scores \\
NC & Normal Cognition \\
NR-VQA & No-Reference Video Quality Assessment \\
SQS & Sequence Quality Score \\
SSL & Self-Supervised Learning \\
\multirow{3}{2.cm}{SSL-V3} & \multirow{3}{6cm}{Self-Supervised Learning-based NR-VQA Combined with ViViT for Video Classification} \\
 & \\ & \\
SSR & Sequence Score Regressor \\
VQA & Video Quality Assessment \\
VQS & Video Quality Score \\
VSR & Video Score Regressor \\
\hline
\end{tabular}
\vspace{-0.8cm}
\label{tab:glssry}
\end{table} 

\end{appendices}

\bibliography{refss}
\end{document}